\documentclass{ecai} 

\usepackage{latexsym}
\usepackage{amssymb}
\usepackage{amsmath}
\usepackage{amsthm}
\usepackage{booktabs}
\usepackage{enumitem}
\usepackage{graphicx}
\usepackage{color}
\usepackage{algorithm}
\usepackage{algorithmic}
\usepackage{subcaption}
\usepackage[british]{babel}
\usepackage{booktabs} 
\usepackage{adjustbox}
\usepackage{multirow}
\usepackage{mathtools} 

\newcommand{\BibTeX}{B\kern-.05em{\sc i\kern-.025em b}\kern-.08em\TeX}

\begin{document}


\begin{frontmatter}




\title{Optimistic Exploration for Risk-Averse Constrained Reinforcement Learning}

\author[A,B]{\fnms{James}~\snm{McCarthy}\thanks{Corresponding Author. Email: james.mccarthy1@ibm.com}}
\author[A]{\fnms{Radu}~\snm{Marinescu}}
\author[A]{\fnms{Elizabeth}~\snm{Daly}}
\author[B]{\fnms{Ivana}~\snm{Dusparic}}

\address[A]{IBM Research Ireland}
\address[B]{Trinity College Dublin}


\begin{abstract}
    Risk-averse Constrained Reinforcement Learning (RaCRL) aims to learn policies that minimise the likelihood of rare and catastrophic constraint violations caused by an environment's inherent randomness. In general, risk-aversion leads to conservative exploration of the environment which typically results in converging to sub-optimal policies that fail to adequately maximise reward or, in some cases, fail to achieve the goal.
    In this paper, we propose an exploration-based approach for RaCRL called Optimistic Risk-averse Actor Critic (ORAC), which constructs an exploratory policy by maximising a local upper confidence bound of the state-action reward value function whilst minimising a local lower confidence bound of the risk-averse state-action cost value function. Specifically, at each step, the weighting assigned to the cost value is increased or decreased if it exceeds or falls below the safety constraint value. This way the policy is encouraged to explore uncertain regions of the environment to discover high reward states whilst still satisfying the safety constraints. Our experimental results demonstrate that the ORAC approach prevents convergence to sub-optimal policies and improves significantly the reward-cost trade-off in various continuous control tasks such as Safety-Gymnasium and a complex building energy management environment CityLearn.

\end{abstract}

\end{frontmatter}

\section{Introduction}
In safety critical domains, it is essential for Reinforcement Learning (RL) systems to adhere to the safety constraints of the environment \citep{garcia_comprehensive_2015,gu_review_2023}. Constrained RL (CRL) has emerged as a framework to address this, by training policies to maximise a reward function whilst satisfying safety constraints over a set of cost functions \citep{achiam_constrained_2017,tessler_reward_2019}. Uncertainty that arises in many real-world scenarios, however, can pose significant challenges to the safety of CRL policies \citep{dulac-arnold_challenges_2021}, as the standard CRL approach is to satisfy constraints in expectation. This leaves CRL policies susceptible to constraint violations caused by the inherent randomness of the environment. Because of this, Risk-averse CRL (RaCRL) has gained attention recently as it aims to minimise the probability of rare but catastrophic constraint violations due to environmental randomness \citep{yang_safetyconstrained_2023,kim2024spectralrisk}. 

In general, however, introducing risk-aversion leads to a more pronounced trade-off between reward and cost, as it induces conservatism in the policy by reducing the set of feasible constraint satisfying policies \cite{yang_risk-aware_2024}. This set of feasible policies will naturally depend on the desired level of risk-aversion but it will also depend on the extent of inherent uncertainty within the environment. \citet{yang_safetyconstrained_2023} propose an off-policy actor-critic RaCRL algorithm called Worst-Case Soft Actor Critic (WCSAC), that makes use of Distributional RL \citep{dabney_implicit_2018,bellemare_distributional_2023} to learn a distributional representation of cost function returns to capture the environment's inherent uncertainty. Using this 
representation, WCSAC calculates the Conditional Value at Risk (CVaR) \citep{rockafellar_optimization_2000} of the policy's worst-case cost returns. Through this, WCSAC learns a policy that, at deployment time, maintains worst-case cost returns below the safety thresholds. 

In this paper, we observe that this conservatism can prevent policies from finding the correct risk-averse solution in a risky Gridworld environment, originally presented in \citep{greenberg_efficient_2022}, where an agent must navigate to a goal block by taking one of two paths -- a short but risky path or a long but safer path. Specifically, we show empirically that, that due to this conservatism, WCSAC under explores the environment and gets stuck in a sub-optimal policy, thus failing to find the risk-averse path to the goal. Subsequently, we apply the principle of Optimism in the Face of Uncertainty (OFU) to the exploration process of the agent and, inspired by \citep{ciosek_better_2019}, we propose an exploration-based approach for risk-averse off-policy constrained RL which we call Optimistic Risk-averse Actor-Critic (ORAC). 

In particular, ORAC constructs an exploratory policy such that at each exploration step it maximises a local upper confidence bound of the state-action reward value function whilst minimising a local lower confidence bound of the risk-averse state-action cost value function. We also develop a weighting scheme for the cost value that increases or decreases at each step should it exceed or fall below the safety constraint value. This way, by leveraging epistemic uncertainty, ORAC bias' the policy's actions towards those that may lead to higher reward states whilst minimising worst-case costs. ORAC encourages the policy to explore uncertain regions of the environment to discover high reward states whilst still encouraging satisfaction of the safety constraint.

We evaluate ORAC in various environments, including the risky Gridworld \citep{greenberg_efficient_2022}, the Safety-Gymnasium's PointGoal1 and PointButton1 environments \citep{ji2023safety}, as well as a complex building energy management environment called CityLearn \citep{vazquez-canteli_citylearn_2019}. Our results demonstrate conclusively that ORAC finds the correct risk-averse path to the goal in the risky Gridworld environment thus avoiding getting stuck in sub-optimal policies. Furthermore, we show considerable improvements in terms of maximising reward whilst minimising cost compared with state-of-the-art risk-averse baselines such as WCSAC in the other environments.



\section{Related Work}
Specification of safety in RL is a persistently challenging problem \citep{garcia_comprehensive_2015,dulac-arnold_challenges_2021}. Constrained RL (CRL) has emerged as an effective framework for maximising a reward objective whilst ensuring satisfaction of constraints over a set of safety objectives  \citep{achiam_constrained_2017,ray_benchmarking_2019,tessler_reward_2019,liu_constrained_2022,zhang_penalized_2022}. Standard CRL is in general, however, neutral to infrequent but catastrophic constraint violations that may arise due to inherent uncertainty within the environment, as safety constraints are satisfied in expectation. \citet{yang_safetyconstrained_2023} aim to mitigate risk in CRL and propose an off-policy CRL approach, extending Soft Actor-Critic to CVaR safety thresholds. The authors propose the use of a distributional value function for the safety critic, specifically the Implicit Quantile Network architecture \citep{dabney_distributional_2018}, allowing for the learning of a expressive representation of the quantile distribution of cost-returns. \citet{kim_trust_2023} proposed a trust-region based algorithm that aims to satisfy mean/standard deviation constraints, reducing uncertainty in constraint satisfaction, but not specifically risk, as mean/standard deviation is not a coherent risk measure. \citet{yang_risk-aware_2024} proposed an on-policy approach using backward value functions, for non-stationary policies, showing strong constraint satisfaction in multiple environments but at a severe reward-cost trade-off. Recently, \citet{kim2024spectralrisk} propose a general approach to linearise spectral risk-measures, such as CVaR, providing strong computational benefits over the non-linear approximation used in \cite{yang_safetyconstrained_2023}. 

Exploration is a widely studied area in standard RL, but is less studied in Constrained RL. \citet{bharadhwaj_conservative_2021} present a conservative exploration approach that resamples actions from a policy if the conservative expected value estimate of the action exceeds the safety threshold. \citet{wachi_safe_2023} present a generalised formulation for safe exploration in constrained RL where the action the agent takes in the environment is interrupted if it is going to violate constraints, and the environment is then reset. This approach is akin to shielding \citep{carr_safe_2023}, in which the agent is prevented from taking an action that leads to constraint violations. These methods, however, rely on strict assumptions such as prior knowledge of the underlying MDP, access to a safe policy or set of actions, and may lead to overly conservative exploration, particularly in complex environments.

We take inspiration from optimistic exploration strategies found in \textit{reward-only} RL literature. In the discrete action space, \citet{keramati_being_2020} aim to optimistically explore the environment to quickly learn a risk-averse policy by forming an optimistic return distribution for action selection. In the continuous action setting, 
\citet{ciosek_better_2019} present an optimistic exploration approach for Actor Critic algorithms such as SAC \citep{haarnoja_soft_2018} and TD3 \citep{fujimoto_addressing_2018} to address conservative exploration that is inherent in these algorithms due to conservative actor updates, inspiring our work. The proposed approach is limited as it only incorporates epistemic uncertainty rather than aleatoric uncertainty which is the primary concern of Risk-Averse RL. \citet{moskovitz_tactical_2022} present an approach that considers aleatoric uncertainty through learning a distributional reward value function, and switch between being optimistic and conservative in the actor updates of the algorithm, rather than through exploration. \citet{liu_ovdexplorer_2024} propose an optimistic exploration strategy that considers both sources of uncertainty. Treating aleatoric uncertainty as a form of risk, the agent is guided away from regions of high aleatoric uncertainty. The authors approach is different to our work in its aims, namely it is focused on risk-neutral maximisation of reward in reward-only MDP settings, whereas our work aims to maximise reward whilst ensuring risk-based satisfaction of cost constraintsr, balancing optimism with explicit constraint satisfaction.


\section{Background}
We begin by providing preliminaries on Constrained Reinforcement Learning CRL, then discuss the extension of CRL to the risk-averse setting and finally we detail how to approximate risk using Distributional RL. 

\subsection{Constrained RL}
A \emph{Constrained Markov Decision Process} \citep{altman_constrained_1999}, is defined by a tuple $(S,A,P,R,C,c,\mu)$, where $S$ and $A$ are the state and action spaces respectively, $P: S \times A \times S \rightarrow \; [0,1]$  is the transition probability function,  $R:S \times A \times S \rightarrow \mathbb{R}$ is the reward function mapping state-action-next state transitions to a real valued scalar, $C:S \times A \times S \rightarrow \mathbb{R}$ is the cost function mapping state-action-next state transitions to a real valued scalar, $c$ is the cost function's corresponding threshold, and $\mu : S \rightarrow [0,1]$ is the starting state distribution.
A stationary policy is a mapping from states to a probability distribution over possible actions, $\pi : S \rightarrow \mathrm{P}(a)$, with $\pi(a|s)$ denoting the probability of taking action $a$ in state $s$. 

In the CMDP setting, the optimal policy is given by:
\begin{equation}\label{eq:CMPD_Opt_Pol}
    \pi^{*} = \arg \max \mathrm{J}_{R}(\pi) \;\; \text{s.t.} \;\; \mathrm{J}_{C}(\pi) \leq c
\end{equation}
where the reward objective $\mathrm{J_{R}}$ is generally taken as the expected infinite horizon discounted return of the reward function $R$, denoted $Q^{\pi}_{R}(s,a)=\mathrm{E}_{\tau \sim \pi}\big[\sum^{\infty}_{t=0}\gamma^{t}R(s_{t},a_{t},s_{t+1})\big]$ and cost objective $\mathrm{J_{C}}$ is the expected infinite horizon discounted return of the cost function $C$, $Q^{\pi}_{C} = \mathrm{E}_{\tau \sim \pi}\big[\sum^{\infty}_{t=0}\gamma^{t}C(s_{t},a_{t},s_{t+1})\big]$. Here, $\tau =\{ s_{0},a_{0},s_{1}, \dots\}$ is a trajectory, and $\tau \sim \pi$ denotes a trajectory induced by policy $\pi$: $s_{0} \sim \mu$, $a_{t}\sim \pi(\cdot|s_{t})$, $s_{t+1}\sim P(\cdot|s_{t},a_{t})$. The primal-dual approach converts Eq. \ref{eq:CMPD_Opt_Pol} into the unconstrained optimisation problem:
\begin{equation}
    \pi^{*} = \min_{\lambda \geq 0} \max_{\pi} \mathrm{E_{\tau \sim \pi}}[Q^{\pi}_{R}(s,a) - \lambda (Q^{\pi}_{C}(s,a) - c)]
\end{equation}

This objective is inherently risk-neutral, however, as it shows equal preference to low cost-returns as it does to rare but catastrophically high cost returns. We now define our Risk-averse Constrained RL (RACRL) problem.



\subsection{Risk Constrained RL}
We define risk as a measure of the likelihood and magnitude of negative effects on our safety objectives, that arise due to inherent uncertainty in the environment \citep{varshney_safety_2017,blokland_concepts_2020,garcia_comprehensive_2015}.
We use the popular Conditional Value at Risk (CVaR) metric \citep{rockafellar_optimization_2000,urpi_riskaverse_2021,greenberg_efficient_2022}, to measure the expected value of the worst $\alpha$ returns of the cost-function:
\begin{equation}
    \text{CVaR}_{\alpha}[X] = \mathrm{E}[X|X \geq F_{X}^{-1}(1-\alpha)]
\end{equation}
where $F_{X}^{-1}$ is the quantile function of the random variable $X$, and $\alpha \in [0,1]$ is our desired risk-level. CVaR can be expressed as a spectral risk measure \cite{acerbi_spectral_2002,kim2024spectralrisk}, expressed in its standard form:
\begin{equation}\label{eq:SpectralRiskFunc}
    \mathcal{R}_{\sigma}(X) \vcentcolon= \int_{0}^{1}F_{X}^{-1}(u)\sigma(u)du
\end{equation}
where $\sigma$ is an increasing function $\sigma \geq 0, \int_{0}^{1}\sigma(u)du = 1$. In the case of CVaR, $\sigma$ is defined as:
\begin{equation}
    \text{CVaR}_{\alpha} \; : \; \sigma(u) = 1_{u\geq \alpha}/(1-\alpha)
\end{equation}

Letting the random variable $Z^{\pi} = \sum_{t=0}^{\infty}\gamma^{t}\mathrm{C}(s_{t},a_{t},s_{s+1})$, denote the discounted sum of cost returns under policy $\pi$, our original risk-neutral CMDP problem, Eq. \ref{eq:CMPD_Opt_Pol}, then becomes:
\begin{equation}\label{eq:RA_CMPD_Opt_Pol}
    \pi^{*} = \arg \max \mathrm{J}_{R}(\pi) \;\; \text{s.t.} \;\; \mathcal{R}_{\sigma}(Z^{\pi})\leq c
\end{equation}

\citet{yang_safetyconstrained_2023} propose Worst-Case Soft Actor-Critic that aims to solve this risk-averse CMDP problem through the following primal-dual optimisation objective:
\begin{equation}\label{eq:WCSACObjective}
    \pi^{*} = \min_{\lambda \geq 0} \max_{\pi} \mathrm{E_{\tau \sim \pi}}[Q^{\pi}_{R}(s,a) - \lambda ( Q_{C,\alpha}^{\pi}(s,a)- c)]
\end{equation}
where $Q_{C,\alpha}^{\pi}(s,a)$ is an approximation of the CVaR of the discounted sum of costs, $\mathcal{R}_{\sigma}(Z^{\pi})$. We next discuss how we learn a parametrised approximation of the distribution $Z^{\pi}$ to approximate $Q_{C,\alpha}^{\pi}(s,a)$

\subsection{Distributional RL}
Recent advances in Distributional RL (DRL) allow for the approximation of the full quantile function for the state-action return distribution, using quantile regression \cite{dabney_distributional_2018}. 
Letting $F_{Z^{\pi}}^{-1}$ denote the quantile function of the random variable $Z^{\pi} = \sum_{t=0}^{\infty}\gamma^{t}\mathrm{C}(s_{t},a_{t},s_{s+1})$, the discounted sum of cost returns under policy $\pi$, we learn a parametrised quantile function of $Z^{\pi}(s,a|\theta)$,  mapping the quantile fraction $k \in [0,1]$ to the quantile value $F_{Z^{\pi}}^{-1}(k)$. 


Using quantile regression, we train the parametrised quantile critic $Z^{\pi}(s,a|\theta)$ by minimising the quantile Huber loss. The Huber loss \citep{huber_robust_1964}, is defined:
\begin{align}
    \mathcal{L}_\kappa(\delta) = \begin{cases}
                                     \frac{1}{2} \delta^2\quad                   & \text{if } |\delta| \le \kappa \\
                                     \kappa(|\delta| - \frac{1}{2}\kappa)\quad   & \text{otherwise}
                                 \end{cases}
\end{align}
The quantile Huber loss is then:
\begin{equation}
    \rho_{k}^{\kappa}(\delta_{i}) = |k - \mathbb{I}\{\delta_{i} <0 \}| \frac{\mathcal{L}_{\kappa}(\delta_{i})}{\kappa}
\end{equation}
where $\delta_{i}$ is the temporal difference error between target distribution $\hat{Z}(s,a)$ and current distribution $Z(s,a)$ defined:
\begin{equation}
    \delta^{t} = c_{t} + \gamma \bigg [ \bar{Z}^{\pi}(s_{t+1},a_{t+1}|\bar{\theta})\bigg] - Z^{\pi}(s_{t},a_{t}|\theta)
\end{equation}

Then using this parametrised quantile distribution we can approximate $\text{CVaR}_{\alpha}(Z^{\pi})$:
\begin{equation}
    Q^{\pi}_{C,\alpha}(s,a) = \frac{1}{1-\alpha}\int_{\alpha}^{1}Z^{\pi}(s,a|\theta)(k)dk
\end{equation}
We further discuss practical implementation details of the approximation of $Q^{\pi}_{C,\alpha}(s,a)$ in the supplemental material.

\section{Optimistic Exploration for Risk-Averse Constrained RL}
In this section we begin by discussing the motivation behind ORAC, we then give a high-level description of ORAC and its aims, finally detailing how it is constructed.

\subsection{Conservative Exploration}
\begin{figure}[t!]
    \centering
    \includegraphics[width=0.65\linewidth]{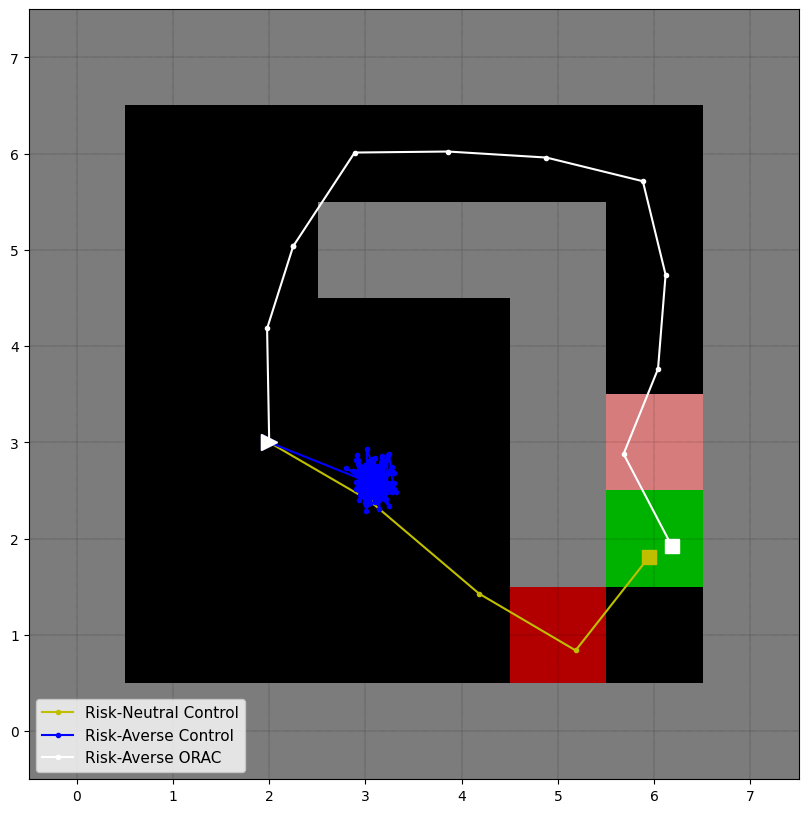}
    \caption{Risk-Averse agent (WCSAC) gets stuck in sub-optimal policy and fails to converge to correct solution. Optimistic Risk-Averse  agent (ORAC) converges to correct risk-averse solution.}
    \label{fig:GuardedMaze}
\vskip 0.2in
\end{figure}

Risk-aversion, in general, leads to more conservative exploration of the environment, as it reduces the set of feasible constraint satisfying policies \citep{yang_safetyconstrained_2023,yang_risk-aware_2024}. We present a motivating example of the effect this conservative exploration can have on a risk-averse CRL agent in a GridWorld environment called GuardedMaze \citep{greenberg_efficient_2022}, shown in Figure \ref{fig:GuardedMaze}. In this environment the agent must navigate from its initial starting position to the green goal block, choosing one of two paths to reach the goal. The shortest path, passing through the red block, leads to higher reward and in the average case incurs less cost. It is, however, more risky as there is a small probability the cost incurred along this path will be far higher than the safety threshold. Instead, the agent may take the long path, passing through the pink block, and receive less reward but incur a fixed cost that satisfies the safety threshold.

We observe that under certain conditions, specifically when the probability of the guard appearing is sufficiently low, risk-averse agents can become stuck in sub-optimal policies, failing to take either path to the goal. The result of this is shown in Figure \ref{fig:GuardedMaze}, where a risk-neutral agent takes the short path and the risk-averse agent gets stuck in a sub-optimal policy, taking neither path. We reason that this is due to conservative exploration of the environment, caused by over-approximation of the cost returns of the long path.

\subsection{Optimistic Exploration}
To address the problem of conservative exploration we extend Worst-Case Soft Actor-Critic (WCSAC) \citep{yang_safetyconstrained_2023}, by constructing a separate exploratory policy following the principle of Optimism in the Face of Uncertainty (OFU), that explores uncertain parts of the state space to discover high reward states, whilst still encouraging satisfaction of the given safety constraints.

In the case of Soft Actor-Critic algorithms, of which WCSAC is itself an extension, actions are sampled from a parametrised action distribution centred around state-dependent mean values, with state-dependent standard deviations. Inspired by \citet{ciosek_better_2019} we propose a method of shifting this mean value towards action values that maximise a locally optimistic upper bound of the reward value-function and minimise a locally optimistic lower bound of the cost value-function.
We begin by defining the learned target policy $\pi_{T}$, which in normal settings is the policy through which the agent explores the environment:
\begin{equation}
    \pi_{T}(a|s) = \mathcal{N}(\mu_{T},\sigma_{T})
\end{equation}
where $\mu_{T} = f_{\mu}(s|\phi)$ is the state-dependent deterministic mean value of policy $\pi_{T}$ and $\sigma_{T} = f_{\sigma}(s|\phi)$ is the state-dependent standard deviation of policy $\pi_{T}$, parametrised by $\phi$. 
Actions are then sampled from this distribution, either side of the mean value $\mu_{T}$, with no preference towards maximising reward or minimising cost. To address this we propose shifting the mean towards a value that specifically maximises or minimises cost, dependent upon whether or not an optimistic lower confidence bound of the cost value for the given state and action is below the safety constraint.

To achieve this, we construct a separate exploration policy, used only to sample actions, at each exploration step with the aim of balancing between maximising a local upper confidence bound of the reward and satisfying a local lower confidence bound of the cost:
\begin{equation}\label{eq:ExpPolArgMax}
    \pi_{E}(a|s) = \arg\max \mathrm{E}_{a\sim \mathcal{N}(\mu,\sigma)}[\hat{Q}_{R}^{\pi}(s,a) - \bar{\lambda} \hat{Q}^{\pi}_{C,\alpha}(s,a)]
\end{equation}
fwhere $\hat{Q}_{R}^{\pi}(s,a)$ is the optimistic upper confidence bound of the reward value function, $\hat{Q}_{C,\alpha}^{\pi}(s,a)$ is the optimistic lower bound of the CVaR of the cost value function, and $\bar{\lambda}$ is the Lagrangian value used to weight the cost value.  We define these terms and how each is calculated in detail, in the following sections.

\subsection{Confidence Bounds of Value Functions}\label{sec:ConfBounds}
To construct our optimistic exploratory policy, we first must define the confidence bounds of both reward and cost value functions.

\paragraph{Reward} We begin by defining the locally optimistic upper confidence bound of our state-action reward value function, following \citet{ciosek_better_2019}:
\begin{equation}\label{eq:OptimQ_r}
    \hat{Q}_{R}^{\pi}(s,a) = \mu_{Q_{R}}(s,a) + \beta_{R}[\sigma_{Q_{R}}(s,a)]
\end{equation}
where $\mu_{Q_{R}}(s,a) = \frac{1}{2}\big[Q^{1}_{R}(s,a) +Q^{2}_{R}(s,a)\big]$ is the mean of the expected state-action reward value, and $\sigma_{Q_{R}}(s,a) = \sqrt{\frac{1}{2}\sum_{i=1}^{2}(Q^{i}_{R}(s,a)-\mu_{Q_{R}}(s,a))}$ is the standard deviation between the two expected state-action reward value.
We control the magnitude of optimism using $\beta_{R} \in \mathbb{R}$. When $\beta_{R}$ is positive $\hat{Q}_{R}(s,a)$ approximates a local upper confidence bound of the state-action reward value function, and when negative it approximates a lower bound. 

\paragraph{Cost} Next we define the locally optimistic lower confidence bound of the state-action cost quantile value function, denoted $\hat{Z}^\pi(s,a)$. We extend WCSAC \citep{yang_safetyconstrained_2023} by learning E independent quantile critics $Z^{\pi}(s,a|\theta_{e})$, that share the same network architecture and are each initialised under different sets of weights. During training, each critic learns the cost returns independently and in parallel using the same collected experience.

To calculate $\hat{Z}^\pi(s,a)$ we first calculate a lower confidence bound of each quantile value across E parametrised quantile functions. Specifically, for each quantile index \(k \in \{1, \dots, K\}\), we construct the lower confidence bound of the \(k\)-th quantile value as:
\begin{equation}\label{eq:Optim_qsa}
    \hat{q}_{C}^{k}(s,a) = \mu_{q_{C}}^{k}(s,a) - \beta_{C}[\sigma_{q_{C}}^{k}(s,a)]
\end{equation}
where $\mu^{k}_{q_{C}}(s,a) = \frac{1}{E}\big[\sum_{e=1}^{E} q^{k}_{C,e}(s,a)\big]$ is the mean value of the $k$-th quantile values drawn from each critic, and $\sigma_{q_{C}}(s,a) = \sqrt{\frac{1}{E}\sum_{e=1}^{E}\big(q^{k}_{C,e}(s,a) -\mu^k_{q_{C}}(s,a)\big)^2}$ is the standard deviation of each $k$-th quantile value from the mean value of the $k$-th quantile value. The locally optimistic lower confidence bound of the state-action cost function $\hat{Z}^{\pi}(s,a)$ is constructed using K optimistic quantiles $q^{\pi}_{c}(s,a)$. Then the locally optimistic CVaR of the cost function can be approximated following\footnote{We discuss extensions of this to other approximations of $\hat{Q}^{\pi}_{C,a}(s,a)$ in the supplemental material.}:
\begin{equation}\label{eq:OptimQsa} 
\hat{Q}^{\pi}_{C,\alpha}(s,a) \approx \frac{1}{1 - \alpha} \sum_{k \in [\alpha, 1]} \hat{q}_{C}^{k}(s,a)
\end{equation} 

This construction allows our exploratory policy to consider two sources of uncertainty, epistemic uncertainty over the set of cost critics, and aleatoric uncertainty captured through the quantile values. It optimistically explores the environment through the optimistic confidence bounds on both reward and cost value functions, whilst importantly maintaining its risk-aversion. 

\begin{algorithm}[t!] 
    \caption{Optimistic Risk-averse Actor Critic (ORAC)}
    \label{alg:ORAC}
    \begin{algorithmic}[1]
        \STATE \textbf{Initialise:} Policy parameters $\phi$, reward critics parameters $\theta_{R_{[1,2]}}$, safety critic parameters \{$\theta_{C}\}_{i=1}^{E}$, target critics $\bar{\theta}_{R_{[1,2]}}$,\{$\bar{\theta}_{C}\}_{i=1}^{E}$, risk-aversion level $\alpha$, Lagrangian multiplier $\lambda$, cost threshold $c$, replay buffer $\mathcal{D}$
        \FOR{each environment step}
        \STATE $\mu_{T}, \sigma_{T} \leftarrow \pi(\cdot|s)$
        \STATE Calculate $\hat{Q}_{UB}(s,a)$ \COMMENT{eq. \ref{eq:OptimQ_r}}
        \STATE Calculate $Q^{C}_{LB}(s,a)$ \COMMENT{eq. \ref{eq:OptimQsa}}
        \STATE Calculate $\bar{\lambda}$ \COMMENT{eq.\ref{eq:AdjLagrangian}}
        \STATE Construct $\mu_{E}$ \COMMENT{eq. \ref{eq:ExpPolBound}}
        \STATE $a_{t} \sim \mathcal{N}(\mu_{E},\sigma_{T})$ \COMMENT{eq. 
 \ref{eq:ExpPolNormDist}}
        \STATE $s_{t+1} \sim P(\cdot|s_{t},a_{t})$
        \STATE Add $\{s_{t},a_{t},r(s_{t},a_{t},s_{t+1}),c(s_{t},a_{t},s_{t+1}),s_{t+1}\}$ to replay buffer $\mathcal{D}$
        \ENDFOR
        \FOR{each gradient step}
        \STATE Update $\phi$, $\theta_{R_{[1,2]}}$, \{$\theta_{C}\}_{i=1}^{E}$,  $\bar{\theta}_{R_{[1,2]}}$,\{$\bar{\theta}_{C}\}_{i=1}^{E}$ Lagrangian multiplier $\lambda$ following standard WCSAC approach
        \ENDFOR
    \end{algorithmic}
\end{algorithm}

\subsection{Policy Construction}
We now detail exactly how the exploration policy $\pi_{E}$ is constructed at each step, using the confidence bounds defined above. We outline the process in Algorithm \ref{alg:ORAC}. Recall that during exploration actions are normally sampled from a target policy distribution $\pi_{T} = \mathcal{N}(\mu_{T},\sigma_{T})$. Our aim is to shift the mean of this distribution towards values that maximise our objective Eq. \ref{eq:ExpPolArgMax}. To do this, we calculate the gradient of our objective Eq. \ref{eq:ExpPolArgMax} at the target policies action $\mu_{T}$:
\begin{equation}\label{eq:ExpPolGrad}
    \nabla_{a}[\hat{Q}_{R}^{\pi}(s,a) - \bar{\lambda} \hat{Q}_{C,\alpha}^{\pi}(s,a)]_{a=\mu_{T}}
\end{equation}
Using automatic differentiation this value is trivial to compute, at minimal computational cost. Then we calculate the change to the target policies action $\mu_{T}$ needed to bias the action towards one that maximises our objective. Following \citet{ciosek_better_2019} we calculate this new mean value such that the KL divergence between the target policy $\pi_{T}$ and the exploratory policy is bounded by $\delta \in \mathbb{R}$. The action mean of the exploratory policy $\mu_{E}$ is calculated by:
\begin{equation}\label{eq:ExpPolBound}
    \mu_{E} = \mu_{T} + \delta \bigg[\frac{\Sigma_{T}\nabla_{a}[\hat{Q}_{R}^{\pi}(s,a) - \bar{\lambda} \hat{Q}_{C,\alpha}^{\pi}(s,a)]_{a=\mu_{T}}}{||\nabla_{a}[\hat{Q}_{R}^{\pi}(s,a) - \bar{\lambda} \hat{Q}_{C,\alpha}^{\pi}(s,a)]_{a=\mu_{T}}||_{\Sigma_{T}}}\bigg]
\end{equation}
Where $\Sigma_{T}$ is the variance of the target policy, calculated by ${\sigma_{T}}^2$, and
$\bar{\lambda}$ is the value used to weight the contribution of the cost-value:
\begin{equation}\label{eq:AdjLagrangian}
    \bar{\lambda} = \lambda - \big[\bar{c} - \hat{Q}^{\pi}_{C,\alpha}(s,a)]
\end{equation}
where $\lambda$ is the Lagrangian value learned throughout training that balances reward and cost in the policy's actor update, and $\bar{c}$ is the cost value constraint. Similarly to \citep{wu_offpolicy_2024} we clip this value as $\max(\bar{\lambda},0)$. Adjusting the learned Lagrangian value in this fashion allows for a reduction in the contribution of the cost-value to the overall gradient Eq. \ref{eq:ExpPolGrad}, if the cost-value is less than the constraint value. This increases the effective weighting of the reward value, shifting actions towards actions that are more reward maximising than cost minimising.

Finally ORAC's exploration policy can be constructed by:
\begin{equation}\label{eq:ExpPolNormDist}
    \pi_{E}(a|s) = \mathcal{N}(\mu_{E},\sigma_{T})
\end{equation}

By biasing the actions of our target policy towards ones that maximise our objective, we are more likely to sample actions that maximise our optimistic objective Eq. \ref{eq:ExpPolArgMax} by exploring uncertain regions of the state space to uncover higher reward states whilst still satisfying constraints.

\begin{figure*}[t!]
    \centering
    \includegraphics[width=0.85\linewidth]{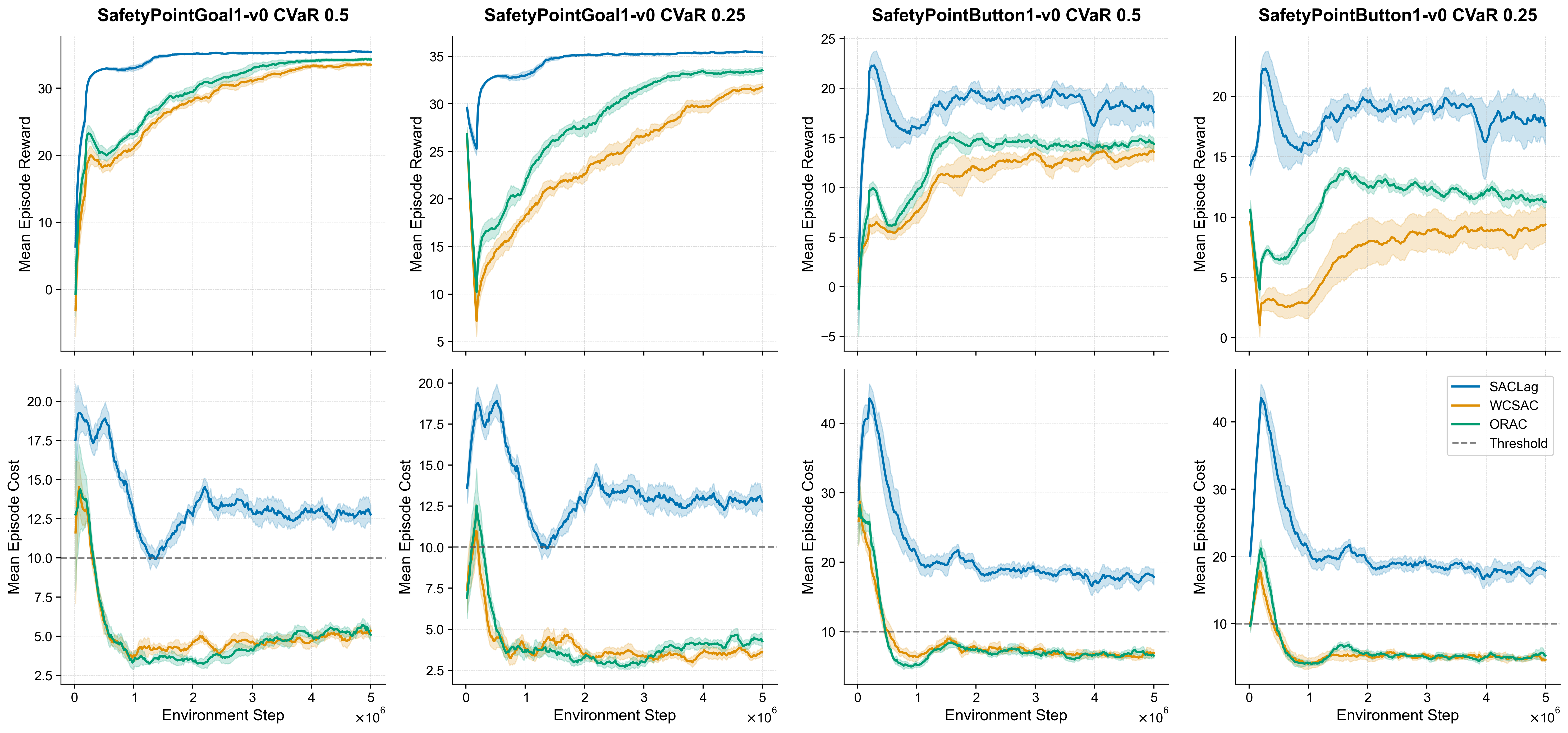}
    \caption{Results of policy evaluations during training of each agent in the Safety Gymnasium environments. The solid line represents the mean of each metric with the shaded area indicating the standard deviation amoung samples, scaled by 0.5 for clarity. Results are obtained by training each agent across 10 seeds. }
    \label{fig:EvalSafeGym}
    \vskip 0.1in
\end{figure*}

\section{Experiments}
In our experiments we seek to evaluate three criteria: (a) can conservatism introduced by risk-aversion lead to agents getting stuck in sub-optimal policies? (b) can optimistic exploration introduced by ORAC prevent agents getting stuck in sub-optimal policies? and (c) does ORAC improve reward-cost trade off by finding policies that achieve higher episodic reward, whilst minimising worst-case costs?

\subsection{Baseline Algorithms}
To evaluate the potential improvements of ORAC in learning risk-averse policies we compare against two baselines. First we compare against a risk-neutral off-policy CRL algorithm Soft Actor-Critic (SAC) Lagrangian, which extends SAC \citep{haarnoja_soft_2018} to the CRL, and aims to satisfy safety constraints in expectation. Next we compare against a state-of-the-art risk-averse off-policy algorithm, Worst-Case SAC (WCSAC) \citep{yang_safetyconstrained_2023}. 

We compare these baseline algorithms against ORAC in three sets of environments, the GuardedMaze Gridworld \citep{greenberg_efficient_2022}, Safety-Gymnasium \citep{ji2023safety} and CityLearn \citep{vazquez-canteli_citylearn_2019}, across 10 seeds for each agent. We note that in complex environments WCSAC can fail to converge to a risk constraint satisfying policy \cite{kim2024spectralrisk}, which we hypothesise is due to the non-linear nature of approximation of risk in the original WCSAC. We therefore extend WCSAC by adopting the approach of \citet{kim2024spectralrisk} to linearise the approximation of risk for Safety Gymnasium and CityLearn environments \footnote{We provide further details of the practical implementation of both the original and linear approximation 
 of risk \cite{kim2024spectralrisk} for WCSAC, comparisons of each approach and ablations of the introduced hyperparameters in supplemental material}. We maintain the original implementation of WCSAC in the GuardedMaze environment as applying the approach of \cite{kim2024spectralrisk} failed to find the solution under any conditions, which we believe is due to extra optimisation challenges posed by this approach.

We detail the hyper-parameters used for each algorithms in each environments, in the supplemental material.  Specifically we use an upper bound value for the reward value function, $\beta_{R}$, of 3.0 in GuardedMaze and 4.0 in Safety Gymnasium and CityLearn. In GuardedMaze we use a larger optimistic lower bound value for the cost value function, $\beta_{C}$, of 2.0, compared to a smaller bound of 1.0 for all other environments. We use an exploration $\delta$ value of 4.0 that we linearly decay to 0 over the course of training. For fairness of comparison we use the same number of cost value functions for both WCSAC and ORAC, 2 in GuardedMaze and 5 in Safety Gymnasium and CityLearn.

\subsection{Optimistic Exploration}
\begin{table}[t!]
\caption{Convergence rate of policies to long path solution in GuardedMaze and average number of steps needed to converge, at different probabilities of a Guard being present. }
\label{tab:GuardedMazeResults}
\begin{center}
\begin{small}
\begin{sc}
\begin{tabular}{lccc}
\toprule
    Agent&Guard Prob.&Success\%&No. Steps\\
         \midrule
         SAC Lag        &0.15 & 0 & - \\
         WCSAC          &0.15 & 60\% & 308000 \\
         ORAC           &0.15 & \textbf{100\%} & \textbf{230000} \\
         \midrule
         SAC Lag        &0.1 & 0 & - \\
         WCSAC          &0.1 & 30\% & 425000 \\
         ORAC           &0.1 & \textbf{80\%} & 263000 \\
    \end{tabular}
\end{sc}
\end{small}
\end{center}
\end{table}

\begin{table*}[t!]
\caption{Results in SafetyGymnasium's PointGoal1 and PointButton1. Safety threshold set to 10 for all environments.}
\label{tab:SafetyGymResults}
\begin{center}
\begin{adjustbox}{width=1\textwidth}
\begin{tabular}{lcccccccc}\toprule
    & \multicolumn{4}{c}{PointGoal1} & \multicolumn{4}{c}{PointButton1}  \\
    \cmidrule(lr){2-5} \cmidrule(lr){6-9} 
    & Reward & Mean Cost & $\text{CVaR}_{0.5}$ & $\text{CVaR}_{0.25}$  & Reward & Mean Cost & $\text{CVaR}_{0.5}$ & $\text{CVaR}_{0.25}$  \\
    \midrule
    SAC Lag                 &35.1$\pm$0.6  &15.5$\pm$5.1           &25.4$\pm$6.4&31.0$\pm$6.9
                            &14.6$\pm$4.7  &19.7$\pm$4.3           &29.4$\pm$6.3&36.1$\pm$8.6 \\ 
                            
    WCSAC-0.5               &33.8$\pm$0.5           &5.0$\pm$1.1             &8.7$\pm$2.0            &11.9$\pm$3.2
                            &14.2$\pm$3.5  &6.7$\pm$2.4   &11.9$\pm$2.8      &16.5$\pm$6.0 \\ 
                            
    ORAC-0.5                &\textbf{34.2$\pm$1.2}           &4.9$\pm$0.9  &8.9$\pm$1.5  &12.4$\pm$2.0
                            &\textbf{13.9$\pm$2.7}             &5.4$\pm$2.1          &9.6$\pm$3.5             &13.2$\pm$5.2 \\
    \midrule
    WCSAC-0.25              &32.1$\pm$1.3          &3.9$\pm$1.5            &7.2$\pm$2.5             &10.6$\pm$5.4
                            &10.1$\pm$3.7           &4.7$\pm$1.1            &8.3$\pm$1.5            &11.4$\pm$2.8\\
    
    ORAC-0.25               &\textbf{32.8$\pm$2.4}           &3.9$\pm$1.1  &7.2$\pm$1.9    &10.0$\pm$2.9
                            &11.2$\pm$3.1   &4.7$\pm$1.8   &8.7$\pm$3.0  &12.7$\pm$4.7\\
    \bottomrule
\end{tabular}
\end{adjustbox}
\end{center}
\end{table*}

To evaluate whether or not ORAC can prevent policies from getting stuck in sub-optimal policies through optimistic exploration, we make use of the GuardedMaze GridWorld environment, originally presented in \citep{greenberg_efficient_2022}, shown in Figure \ref{fig:GuardedMaze}. The objective of the agent is to navigate from a random initial state in the lower left quarter of the grid, to the green goal block. The feature space consists of the agent's $xy$ co-ordinates in the grid and its action space is a continuous two-dimensional action space controlling its movement along the x and y axes. Adapting the environment to the CRL setting we add cost blocks along each path to the goal. At each step, the agent receives a reward of -1, for the first 32 steps of an episode and 0 thereafter, and +16 if it reaches the goal block at which point the episode is terminated. Along the short and most rewarding path, the agent passes through a guarded block (red block), where a guard will be present with a probability $p$, at each episode initialisation. The guard is not part of the agents observed feature space, but only part of the environment. If the guard is not present, the cost incurred by passing through the guarded block is 2, but if the guard is present the cost is much higher at 20. The longer path to the goal is less rewarding as the agent must take more steps, and passing through the cost block (pink block) incurs a fixed cost of 4. We add a bonus of +1 in the cell in the upper right corner of the grid to help incentivise exploration of this path as we found in early experiments agents converged to a policy that got stuck in the upper right corner of the environment and didn't sufficiently explore the long path. If the agent fails to reach the goal block, the episode will be truncated at 100 steps and the environment will reset, beginning a new episode.

We evaluate each algorithm under two guard probability settings, 15\% and 10\%. Each algorithm is trained across 10 seeds, for 500K timesteps. We set the cost threshold to 5 and the risk-aversion level, $\alpha$, to 0.05, meaning the risk-averse algorithms are trained to satisfy the threshold of 5 in 95\% of the episodes. Under these conditions, the short path satisfies the cost threshold in the average case, but exceeds the threshold in the worst 5\% of cases, making the long path the less rewarding but risk optimal path. 

We detail the results of these evaluations in Table \ref{tab:GuardedMazeResults}. The risk-neutral agent behaves as expected and converges to the short path under both conditions, for each seed. The risk-averse baseline WCSAC, however, only converges to the risk-averse long path 60\% of the time under the 0.15 probability setting and even less so under the 0.1\% at just 30\%, indicating WCSAC insufficiently explores the environment to find the long path. In contrast, ORAC converges to the risk-averse long path 100\% of the time under the 0.15 probability setting and 80\% of the time in the 0.1 probability setting, demonstrating clearly the improved exploration of ORAC. We measure the number of steps taken for the algorithms to converge to the long path solution, measuring failed solutions as 500K steps. We note that ORAC leads to faster convergence to the long path under both Guard probability settings, indicating ORAC leads to more sample efficient exploration over the baseline.

\subsection{Cost-Reward Trade-off Evaluation}

\begin{table*}[t!]
\caption{Results CityLearn Environment. Safety threshold is 720}
\label{tab:CityLearnResults}

\begin{center}
\begin{small}
\begin{sc}
\begin{adjustbox}{width=1\textwidth}
    \begin{tabular}{lccccc}
        \toprule
        & Reward & Mean Cost & $\text{CVaR}_{0.05}$ &Unserved Energy in Outage (\%)&Outage Temp. Violations (\%)\\
        \midrule
        SAC Lag           &-605.9$\pm$8.3       &772.6$\pm$15.2            &861.2$\pm$27.0                        &\textbf{0.61$\pm$0.01}&0.73$\pm$0.01\\ 
        WCSAC-0.05        &-1417.6$\pm$118.9    &\textbf{623.3$\pm$39.1}    &\textbf{699.7$\pm$38.7}         &0.66$\pm$0.03 &0.70$\pm$0.04 \\
        ORAC-0.05         &\textbf{-1299.3$\pm$57.4}    &629.6$\pm$17.6             &702.8$\pm$25.9               &0.63$\pm$0.02 &\textbf{0.65$\pm$0.03}\\
    \end{tabular}
    \end{adjustbox}
\end{sc}
\end{small}
\end{center}
\end{table*}
To assess whether or not ORAC improves the reward-cost trade-off over the baseline risk-averse WCSAC by converging to higher reward policies, we evaluate the algorithms in the Safety-Gymnasium environments PointGoal1 and PointButton1 \citep{ji2023safety}, and with complex building energy management simulation environment, CityLearn \citep{vazquez-canteli_citylearn_2019}. 

\paragraph{SafetyGym} The Safety-Gymnasium environments, presented by \citep{ji2023safety}, act as the standard baseline environments for constrained RL and present challenging environments in which agents balance the reward-cost trade-off. In the PointGoal1 and PointButton1 environments the agent must navigate from an initial starting position to a goal objective, whilst avoiding obstacles. In PointGoal1, the objective is to reach a goal position whilst avoiding static obstacles. In PointButton1 the objective is to press one of a set of randomly positioned buttons, whilst avoiding both static and moving obstacles, making it the more challenging environment. The observation spaces consist of a large number of features representing lidar readings of the agent, indicating the distance to other objects. The reward function is a dense function measuring the absolute distance towards the goal and a constant for reaching the goal location or pushing the correct button. Unsafe behaviour of colliding with an obstacle, entering a static obstacle zone or pushing the wrong button return a cost of 1, whilst safe behaviour returns a cost of 0.

We set the cost constraint to 10 for each episode and truncate the episodes to a max length of 400 steps, adapting changes described in \citep{liu_constrained_2022, wu_offpolicy_2024} to speed up training of the agents. Specifically, the simulation timesteps of the environment are increased, but otherwise the environments are unchanged. We assess WCSAC and ORAC under two risk-averse settings, $\text{CVaR}_{0.5}$ and $\text{CVaR}_{0.25}$, to observe the behaviour of ORAC under increasing risk-aversion. We train the agents for 5 million timesteps across 10 seeds. We plot the evaluations of the agents at different points throughout training for each environment at each risk level in Figure \ref{fig:EvalSafeGym}. We then evaluate the fully trained agents in 100 episodes per seed and report the mean episodic reward and cost, and the CVaR episodic cost at each level of risk-aversion. These results are summarised in Table \ref{tab:SafetyGymResults}.

Figure \ref{fig:EvalSafeGym} shows the evaluation runs of the agents in the two environments. ORAC is consistently higher in terms of Mean Episodic Reward throughout training in all environments, with WCSAC eventually converging to a similar reward in PointButton1 at the CvaR-0.5 risk level. Early in training in PointGoal1 ORAC achieves lower cost than WCSAC, whilst maintaining higher reward performance. This highlights ORAC more quickly finds a policy that exploits reward whilst minimising cost. 

Table \ref{tab:SafetyGymResults} summarises the results of post training evaluations. In PointGoal1, ORAC improves on the performance of WCSAC, by achieving a higher reward whilst satisfying the CVaR cost thresholds. In PointButton1, we note that WCSAC marginally fails to satisfy the constraints at all risk-levels. ORAC on the other hand, satisfies the constraints at the 0.5 risk-level, but similarly to WCSAC marginally violates the constraint at the 0.25 risk-level. At the 0.5 risk-level ORAC achieves a slightly lower reward but satisfies the cost constraint, whilst at the 0.25 risk-level ORAC achieves a significantly higher reward and a slightly higher cost.

In the Safety Gymnasium environments we observe a improvement in performance of ORAC over WCSAC. In PointGoal it improves upon the reward of WCSAC whilst satisfying CVaR cost constraints. And in PointButton1, ORAC satisfies the CVaR-0.5 cost constraint whilst WCSAC does not. 

\paragraph{CityLearn}
In the CityLearn environment \citep{vazquez-canteli_citylearn_2019} the agent is tasked with maintaining the thermal comfort of a building, whilst minimising the electricity demand the building places on the grid.
Power-outages occur in the environment randomly and infrequently, cutting the supply of electricity from the grid to each building, and can last for a number of hours. During regular operations the agent uses energy supplied from the grid to control the temperature of the building but during outages it must rely on either solar energy or energy stored in its electrical battery. A risk-averse policy should learn to prepare for power-outages by storing electricity in the battery. Whilst the impact of the outage is dependent on the policy, the likelihood of it occurring is completely independent, therefore, this environment will act a strong test for the efficacy of ORAC in environments where the agent has no control over whether or not it experiences a rare-event, only control over how to minimise the impact of said rare-event. 

In our experiments we represent the goal of reducing the building's demand on the energy grid through a reward function and the goal of maintaining the indoor temperature through a cost function and constraint limit, representing a comfort band around the desired temperature, which is set to 1 degree at every timestep. Each episode runs for a total of 720 timesteps, representing 1 month. We fix the probability of a power outage at approximately 10\% at each timestep, which may appear high but outages occur for consecutive timesteps meaning their impact varies, depending on a number of factors. We train our agents for 3 million timesteps at a risk-level of 0.05, across 10 seeds. We measure the mean episodic reward and cost, and the $\text{CVaR}_{0.05}$ episodic costs. We include two metrics specific to CityLearn, \textit{Outage Temperature Violations} indicating the proportion of timesteps that the agent violates the temperature constraint during an outage, and \textit{Unserved Energy during Outages} indicating the proportion of energy demand that is unmet by the agent specifically during an outage. These metrics highlight the agents performance specifically during outages as the reward and cost metrics are calculated during both regular operation and during outages.

Table \ref{tab:CityLearnResults} shows the results of our evaluations. As expected, the risk-neutral basline, SAC Lag, achieves the highest episodic reward, but returns the highest mean and CVaR cost. The risk-averse baseline, WCSAC, achieves the lowest cost returns but the lowest mean episodic reward. ORAC improves significantly on the mean episodic reward over WCSAC, but returns slightly higher mean and CVaR costs, still satisfying the threshold. ORAC significantly improves upon both baselines in terms of Outage Temperature Violations and achieves an Unserved Energy during Outages score close to SACLag, which indicates better exploitation of reward during outages than WCSAC. We reason that these results along with the increased cost shown by ORAC may indicate a qualitative change in the policy by focusing more on maintaining temperature demands and energy supply during outages, rather than during regular hours.

\section{Discussion}
In this work we presented ORAC, an optimistic exploration approach to Risk-averse CRL for Lagrangian based off-policy actor-critic algorithms. Specifically, ORAC constructs an optimistic exploratory policy at each timestep, that maximises an upper confidence bound of the state-action reward value function whilst minimising a local lower confidence bound of the risk-averse state-action cost value function. ORAC is encouraged to explore uncertain regions of the environment, to discover high reward policies whilst aiming to still satisfy safety constraints. We demonstrated ORAC, by improved exploration, prevents a risk-averse baseline, WCSAC, from converging to sub-optimal policies that fail to find the correct solution in a risky GridWorld environment. We further demonstrated improvements in Safety-Gymnasium environments and in a complex building energy management environment, CityLearn, where ORAC improved reward maximisation whilst minimising cost, over the baseline.

Similarly to the underlying algorithm, WCSAC, however, ORAC does not guarantee safe exploration but instead aims for safe policies at deployment. In future work we aim to investigate an exploration approach that adaptively explores the environment switching between optimism and conservatism based on constraint satisfaction, to improve the safety during training. Further future work could look at most robust estimates of epistemic uncertainty and potentially dynamic confidence bounds, as ORAC makes use of fixed hyperparameter values for the confidence bounds.

\ack{The author would like to acknowledge the support of the IBM Pre-Doc Programme, part-funded by IDA Training Grant 215544.}

\bibliography{references}

\begin{thebibliography}{36}
\providecommand{\natexlab}[1]{#1}
\providecommand{\url}[1]{\texttt{#1}}
\expandafter\ifx\csname urlstyle\endcsname\relax
  \providecommand{\doi}[1]{doi: #1}\else
  \providecommand{\doi}{doi: \begingroup \urlstyle{rm}\Url}\fi

\bibitem[Acerbi(2002)]{acerbi_spectral_2002}
C.~Acerbi.
\newblock Spectral measures of risk: {A} coherent representation of subjective risk aversion.
\newblock \emph{Journal of Banking \& Finance}, 26\penalty0 (7):\penalty0 1505--1518, July 2002.
\newblock ISSN 03784266.
\newblock \doi{10.1016/S0378-4266(02)00281-9}.
\newblock URL \url{https://linkinghub.elsevier.com/retrieve/pii/S0378426602002819}.

\bibitem[Achiam et~al.(2017)Achiam, Held, Tamar, and Abbeel]{achiam_constrained_2017}
J.~Achiam, D.~Held, A.~Tamar, and P.~Abbeel.
\newblock Constrained {Policy} {Optimization}.
\newblock In \emph{Proceedings of the 34th {International} {Conference} on {Machine} {Learning}}, pages 22--31. PMLR, July 2017.
\newblock URL \url{https://proceedings.mlr.press/v70/achiam17a.html}.
\newblock ISSN: 2640-3498.

\bibitem[Altman(1999)]{altman_constrained_1999}
E.~Altman.
\newblock \emph{Constrained {Markov} {Decision} {Processes}: {Stochastic} {Modeling}}.
\newblock Routledge, Boca Raton, 1 edition, 1999.
\newblock ISBN 978-1-315-14022-3.
\newblock \doi{10.1201/9781315140223}.
\newblock URL \url{https://www.taylorfrancis.com/books/9781315140223}.

\bibitem[Bellemare et~al.(2023)Bellemare, Dabney, and Rowland]{bellemare_distributional_2023}
M.~G. Bellemare, W.~Dabney, and M.~Rowland.
\newblock \emph{Distributional reinforcement learning}.
\newblock The MIT Press, Cambridge, Massachusetts, 2023.
\newblock ISBN 978-0-262-37402-6.
\newblock OCLC: 1350431122.

\bibitem[Bharadhwaj et~al.(2021)Bharadhwaj, Kumar, Rhinehart, Levine, Shkurti, and Garg]{bharadhwaj_conservative_2021}
H.~Bharadhwaj, A.~Kumar, N.~Rhinehart, S.~Levine, F.~Shkurti, and A.~Garg.
\newblock Conservative {Safety} {Critics} for {Exploration}, Apr. 2021.
\newblock URL \url{http://arxiv.org/abs/2010.14497}.
\newblock arXiv:2010.14497 [cs].

\bibitem[Blokland and Reniers(2020)]{blokland_concepts_2020}
P.~J. Blokland and G.~L. Reniers.
\newblock The {Concepts} of {Risk}, {Safety}, and {Security}: {A} {Fundamental} {Exploration} and {Understanding} of {Similarities} and {Differences}.
\newblock In C.~Bieder and K.~Pettersen~Gould, editors, \emph{The {Coupling} of {Safety} and {Security}}, pages 9--16. Springer International Publishing, Cham, 2020.
\newblock ISBN 978-3-030-47228-3 978-3-030-47229-0.
\newblock \doi{10.1007/978-3-030-47229-0_2}.
\newblock URL \url{http://link.springer.com/10.1007/978-3-030-47229-0_2}.
\newblock Series Title: SpringerBriefs in Applied Sciences and Technology.

\bibitem[Carr et~al.(2023)Carr, Jansen, Junges, and Topcu]{carr_safe_2023}
S.~Carr, N.~Jansen, S.~Junges, and U.~Topcu.
\newblock Safe {Reinforcement} {Learning} via {Shielding} under {Partial} {Observability}.
\newblock \emph{Proceedings of the AAAI Conference on Artificial Intelligence}, 37\penalty0 (12):\penalty0 14748--14756, June 2023.
\newblock ISSN 2374-3468, 2159-5399.
\newblock \doi{10.1609/aaai.v37i12.26723}.
\newblock URL \url{https://ojs.aaai.org/index.php/AAAI/article/view/26723}.

\bibitem[Ciosek et~al.(2019)Ciosek, Vuong, Loftin, and Hofmann]{ciosek_better_2019}
K.~Ciosek, Q.~Vuong, R.~Loftin, and K.~Hofmann.
\newblock Better {Exploration} with {Optimistic} {Actor} {Critic}.
\newblock In \emph{Advances in {Neural} {Information} {Processing} {Systems}}, volume~32. Curran Associates, Inc., 2019.
\newblock URL \url{https://proceedings.neurips.cc/paper_files/paper/2019/hash/a34bacf839b923770b2c360eefa26748-Abstract.html}.

\bibitem[Dabney et~al.(2018{\natexlab{a}})Dabney, Ostrovski, Silver, and Munos]{dabney_implicit_2018}
W.~Dabney, G.~Ostrovski, D.~Silver, and R.~Munos.
\newblock Implicit {Quantile} {Networks} for {Distributional} {Reinforcement} {Learning}.
\newblock In \emph{Proceedings of the 35th {International} {Conference} on {Machine} {Learning}}, pages 1096--1105. PMLR, July 2018{\natexlab{a}}.
\newblock URL \url{https://proceedings.mlr.press/v80/dabney18a.html}.
\newblock ISSN: 2640-3498.

\bibitem[Dabney et~al.(2018{\natexlab{b}})Dabney, Rowland, Bellemare, and Munos]{dabney_distributional_2018}
W.~Dabney, M.~Rowland, M.~Bellemare, and R.~Munos.
\newblock Distributional {Reinforcement} {Learning} {With} {Quantile} {Regression}.
\newblock \emph{Proceedings of the AAAI Conference on Artificial Intelligence}, 32\penalty0 (1), Apr. 2018{\natexlab{b}}.
\newblock ISSN 2374-3468, 2159-5399.
\newblock \doi{10.1609/aaai.v32i1.11791}.
\newblock URL \url{https://ojs.aaai.org/index.php/AAAI/article/view/11791}.

\bibitem[Dulac-Arnold et~al.(2021)Dulac-Arnold, Levine, Mankowitz, Li, Paduraru, Gowal, and Hester]{dulac-arnold_challenges_2021}
G.~Dulac-Arnold, N.~Levine, D.~J. Mankowitz, J.~Li, C.~Paduraru, S.~Gowal, and T.~Hester.
\newblock Challenges of real-world reinforcement learning: definitions, benchmarks and analysis.
\newblock \emph{Machine Learning}, 110\penalty0 (9):\penalty0 2419--2468, Sept. 2021.
\newblock ISSN 0885-6125, 1573-0565.
\newblock \doi{10.1007/s10994-021-05961-4}.
\newblock URL \url{https://link.springer.com/10.1007/s10994-021-05961-4}.

\bibitem[Fujimoto et~al.(2018)Fujimoto, Hoof, and Meger]{fujimoto_addressing_2018}
S.~Fujimoto, H.~Hoof, and D.~Meger.
\newblock Addressing {Function} {Approximation} {Error} in {Actor}-{Critic} {Methods}.
\newblock In \emph{Proceedings of the 35th {International} {Conference} on {Machine} {Learning}}, pages 1587--1596. PMLR, July 2018.
\newblock URL \url{https://proceedings.mlr.press/v80/fujimoto18a.html}.
\newblock ISSN: 2640-3498.

\bibitem[García and Fernández(2015)]{garcia_comprehensive_2015}
J.~García and F.~Fernández.
\newblock A {Comprehensive} {Survey} on {Safe} {Reinforcement} {Learning}.
\newblock \emph{Journal of Machine Learning Research}, 16\penalty0 (42):\penalty0 1437--1480, 2015.
\newblock ISSN 1533-7928.
\newblock URL \url{http://jmlr.org/papers/v16/garcia15a.html}.

\bibitem[Greenberg et~al.(2022)Greenberg, Chow, Ghavamzadeh, and Mannor]{greenberg_efficient_2022}
I.~Greenberg, Y.~Chow, M.~Ghavamzadeh, and S.~Mannor.
\newblock Efficient risk-averse reinforcement learning.
\newblock \emph{Advances in Neural Information Processing Systems}, 35:\penalty0 32639--32652, 2022.

\bibitem[Gu et~al.(2023)Gu, Yang, Du, Chen, Walter, Wang, Yang, and Knoll]{gu_review_2023}
S.~Gu, L.~Yang, Y.~Du, G.~Chen, F.~Walter, J.~Wang, Y.~Yang, and A.~Knoll.
\newblock A {Review} of {Safe} {Reinforcement} {Learning}: {Methods}, {Theory} and {Applications}, Feb. 2023.
\newblock URL \url{http://arxiv.org/abs/2205.10330}.
\newblock arXiv:2205.10330 [cs].

\bibitem[Haarnoja et~al.(2018)Haarnoja, Zhou, Abbeel, and Levine]{haarnoja_soft_2018}
T.~Haarnoja, A.~Zhou, P.~Abbeel, and S.~Levine.
\newblock Soft {Actor}-{Critic}: {Off}-{Policy} {Maximum} {Entropy} {Deep} {Reinforcement} {Learning} with a {Stochastic} {Actor}.
\newblock In \emph{Proceedings of the 35th {International} {Conference} on {Machine} {Learning}}, pages 1861--1870. PMLR, July 2018.
\newblock URL \url{https://proceedings.mlr.press/v80/haarnoja18b.html}.
\newblock ISSN: 2640-3498.

\bibitem[Huber(1964)]{huber_robust_1964}
P.~J. Huber.
\newblock Robust estimation of a location parameter.
\newblock \emph{The Annals of Mathematical Statistics}, 35\penalty0 (1):\penalty0 73 -- 101, 1964.
\newblock \doi{10.1214/aoms/1177703732}.
\newblock URL \url{https://doi.org/10.1214/aoms/1177703732}.
\newblock Publisher: Institute of Mathematical Statistics.

\bibitem[Ji et~al.(2023)Ji, Zhang, Zhou, Pan, Huang, Sun, Geng, Zhong, Dai, and Yang]{ji2023safety}
J.~Ji, B.~Zhang, J.~Zhou, X.~Pan, W.~Huang, R.~Sun, Y.~Geng, Y.~Zhong, J.~Dai, and Y.~Yang.
\newblock Safety gymnasium: a unified safe reinforcement learning benchmark.
\newblock In \emph{Thirty-seventh conference on neural information processing systems datasets and benchmarks track}, 2023.
\newblock URL \url{https://openreview.net/forum?id=WZmlxIuIGR}.

\bibitem[Keramati et~al.(2020)Keramati, Dann, Tamkin, and Brunskill]{keramati_being_2020}
R.~Keramati, C.~Dann, A.~Tamkin, and E.~Brunskill.
\newblock Being {Optimistic} to {Be} {Conservative}: {Quickly} {Learning} a {CVaR} {Policy}.
\newblock \emph{Proceedings of the AAAI Conference on Artificial Intelligence}, 34\penalty0 (04):\penalty0 4436--4443, Apr. 2020.
\newblock ISSN 2374-3468, 2159-5399.
\newblock \doi{10.1609/aaai.v34i04.5870}.
\newblock URL \url{https://ojs.aaai.org/index.php/AAAI/article/view/5870}.

\bibitem[Kim et~al.(2023)Kim, Lee, and Oh]{kim_trust_2023}
D.~Kim, K.~Lee, and S.~Oh.
\newblock Trust {Region}-{Based} {Safe} {Distributional} {Reinforcement} {Learning} for {Multiple} {Constraints}.
\newblock \emph{Advances in Neural Information Processing Systems}, 36:\penalty0 19908--19939, Dec. 2023.
\newblock URL \url{https://proceedings.neurips.cc/paper_files/paper/2023/hash/3f20f2b0315c72201e23512fdbd1ee91-Abstract-Conference.html}.

\bibitem[Kim et~al.(2024)Kim, Cho, Han, Chung, Lee, and Oh]{kim2024spectralrisk}
D.~Kim, T.~Cho, S.~Han, H.~Chung, K.~Lee, and S.~Oh.
\newblock Spectral-risk safe reinforcement learning with convergence guarantees.
\newblock In \emph{The thirty-eighth annual conference on neural information processing systems}, 2024.
\newblock URL \url{https://openreview.net/forum?id=9JFSJitKC0}.

\bibitem[Liu et~al.(2024)Liu, Wang, Zheng, Hao, Bai, Ye, Wang, Piao, and Sun]{liu_ovdexplorer_2024}
J.~Liu, Z.~Wang, Y.~Zheng, J.~Hao, C.~Bai, J.~Ye, Z.~Wang, H.~Piao, and Y.~Sun.
\newblock {OVD}-{Explorer}: {Optimism} {Should} {Not} {Be} the {Sole} {Pursuit} of {Exploration} in {Noisy} {Environments}.
\newblock \emph{Proceedings of the AAAI Conference on Artificial Intelligence}, 38\penalty0 (12):\penalty0 13954--13962, Mar. 2024.
\newblock ISSN 2374-3468, 2159-5399.
\newblock \doi{10.1609/aaai.v38i12.29303}.
\newblock URL \url{https://ojs.aaai.org/index.php/AAAI/article/view/29303}.

\bibitem[Liu et~al.(2022)Liu, Cen, Isenbaev, Liu, Wu, Li, and Zhao]{liu_constrained_2022}
Z.~Liu, Z.~Cen, V.~Isenbaev, W.~Liu, S.~Wu, B.~Li, and D.~Zhao.
\newblock Constrained {Variational} {Policy} {Optimization} for {Safe} {Reinforcement} {Learning}.
\newblock In \emph{Proceedings of the 39th {International} {Conference} on {Machine} {Learning}}, pages 13644--13668. PMLR, June 2022.
\newblock URL \url{https://proceedings.mlr.press/v162/liu22b.html}.
\newblock ISSN: 2640-3498.

\bibitem[Ma et~al.(2020)Ma, Xia, Zhou, Yang, and Zhao]{ma_dsac_2020}
X.~Ma, L.~Xia, Z.~Zhou, J.~Yang, and Q.~Zhao.
\newblock {DSAC}: {Distributional} {Soft} {Actor} {Critic} for {Risk}-{Sensitive} {Reinforcement} {Learning}, June 2020.
\newblock URL \url{http://arxiv.org/abs/2004.14547}.
\newblock arXiv:2004.14547 [cs].

\bibitem[Moskovitz et~al.(2022)Moskovitz, Parker-Holder, Pacchiano, Arbel, and Jordan]{moskovitz_tactical_2022}
T.~Moskovitz, J.~Parker-Holder, A.~Pacchiano, M.~Arbel, and M.~I. Jordan.
\newblock Tactical {Optimism} and {Pessimism} for {Deep} {Reinforcement} {Learning}, Apr. 2022.
\newblock URL \url{http://arxiv.org/abs/2102.03765}.
\newblock arXiv:2102.03765.

\bibitem[Ray et~al.(2019)Ray, Achiam, and Amodei]{ray_benchmarking_2019}
A.~Ray, J.~Achiam, and D.~Amodei.
\newblock Benchmarking {Safe} {Exploration} in {Deep} {Reinforcement} {Learning}.
\newblock page~25, 2019.

\bibitem[Rockafellar and Uryasev(2000)]{rockafellar_optimization_2000}
R.~T. Rockafellar and S.~Uryasev.
\newblock Optimization of conditional value-at-risk.
\newblock \emph{The Journal of Risk}, 2\penalty0 (3):\penalty0 21--41, 2000.
\newblock ISSN 14651211.
\newblock \doi{10.21314/JOR.2000.038}.
\newblock URL \url{http://www.risk.net/journal-of-risk/technical-paper/2161159/optimization-conditional-value-risk}.

\bibitem[Tessler et~al.(2019)Tessler, Mankowitz, and Mannor]{tessler_reward_2019}
C.~Tessler, D.~J. Mankowitz, and S.~Mannor.
\newblock Reward {Constrained} {Policy} {Optimization}.
\newblock 2019.

\bibitem[Urpí et~al.(2021)Urpí, Curi, and Krause]{urpi_riskaverse_2021}
N.~A. Urpí, S.~Curi, and A.~Krause.
\newblock Risk-{Averse} {Offline} {Reinforcement} {Learning}.
\newblock 2021.
\newblock URL \url{https://openreview.net/forum?id=TBIzh9b5eaz}.

\bibitem[Varshney and Alemzadeh(2017)]{varshney_safety_2017}
K.~R. Varshney and H.~Alemzadeh.
\newblock On the {Safety} of {Machine} {Learning}: {Cyber}-{Physical} {Systems}, {Decision} {Sciences}, and {Data} {Products}.
\newblock \emph{Big Data}, 5\penalty0 (3):\penalty0 246--255, Sept. 2017.
\newblock ISSN 2167-6461, 2167-647X.
\newblock \doi{10.1089/big.2016.0051}.
\newblock URL \url{http://www.liebertpub.com/doi/10.1089/big.2016.0051}.

\bibitem[Vázquez-Canteli et~al.(2019)Vázquez-Canteli, Kämpf, Henze, and Nagy]{vazquez-canteli_citylearn_2019}
J.~R. Vázquez-Canteli, J.~Kämpf, G.~Henze, and Z.~Nagy.
\newblock {CityLearn} v1.0: {An} {OpenAI} gym environment for demand response with deep reinforcement learning.
\newblock In \emph{Proceedings of the 6th {ACM} international conference on systems for energy-efficient buildings, cities, and transportation}, {BuildSys} '19, pages 356--357, New York, NY, USA, 2019. Association for Computing Machinery.
\newblock ISBN 978-1-4503-7005-9.
\newblock \doi{10.1145/3360322.3360998}.
\newblock URL \url{https://doi.org/10.1145/3360322.3360998}.
\newblock Number of pages: 2 Place: New York, NY, USA.

\bibitem[Wachi et~al.(2023)Wachi, Hashimoto, Shen, and Hashimoto]{wachi_safe_2023}
A.~Wachi, W.~Hashimoto, X.~Shen, and K.~Hashimoto.
\newblock Safe {Exploration} in {Reinforcement} {Learning}: {A} {Generalized} {Formulation} and {Algorithms}, Oct. 2023.
\newblock URL \url{http://arxiv.org/abs/2310.03225}.
\newblock arXiv:2310.03225 [cs].

\bibitem[Wu et~al.(2024)Wu, Tang, Lin, Yu, Mao, Xie, Wang, and Wang]{wu_offpolicy_2024}
Z.~Wu, B.~Tang, Q.~Lin, C.~Yu, S.~Mao, Q.~Xie, X.~Wang, and D.~Wang.
\newblock Off-policy primal-dual safe reinforcement learning.
\newblock In \emph{The twelfth international conference on learning representations}, 2024.
\newblock URL \url{https://openreview.net/forum?id=vy42bYs1Wo}.

\bibitem[Yang et~al.(2023)Yang, Simão, Tindemans, and Spaan]{yang_safetyconstrained_2023}
Q.~Yang, T.~D. Simão, S.~H. Tindemans, and M.~T.~J. Spaan.
\newblock Safety-constrained reinforcement learning with a distributional safety critic.
\newblock \emph{Machine Learning}, 112\penalty0 (3):\penalty0 859--887, Mar. 2023.
\newblock ISSN 0885-6125, 1573-0565.
\newblock \doi{10.1007/s10994-022-06187-8}.
\newblock URL \url{https://link.springer.com/10.1007/s10994-022-06187-8}.

\bibitem[Yang et~al.(2024)Yang, Jin, Tang, and Fan]{yang_risk-aware_2024}
Z.~Yang, H.~Jin, Y.~Tang, and G.~Fan.
\newblock Risk-aware constrained reinforcement learning with non-stationary policies.
\newblock In \emph{Proceedings of the 23rd international conference on autonomous agents and multiagent systems}, Aamas '24, pages 2029--2037, Richland, SC, 2024. International Foundation for Autonomous Agents and Multiagent Systems.
\newblock ISBN 9798400704864.
\newblock Number of pages: 9 Place: Auckland, New Zealand.

\bibitem[Zhang et~al.(2022)Zhang, Shen, Yang, Chen, Wang, Yuan, and Tao]{zhang_penalized_2022}
L.~Zhang, L.~Shen, L.~Yang, S.~Chen, X.~Wang, B.~Yuan, and D.~Tao.
\newblock Penalized {Proximal} {Policy} {Optimization} for {Safe} {Reinforcement} {Learning}.
\newblock In \emph{Proceedings of the {Thirty}-{First} {International} {Joint} {Conference} on {Artificial} {Intelligence}}, pages 3744--3750, Vienna, Austria, July 2022. International Joint Conferences on Artificial Intelligence Organization.
\newblock ISBN 978-1-956792-00-3.
\newblock \doi{10.24963/ijcai.2022/520}.
\newblock URL \url{https://www.ijcai.org/proceedings/2022/520}.

\end{thebibliography}

\newpage
\onecolumn

\appendix
\section{Hyper-Parameters}
In this section we outline the hyper-parmeters used across environments. Firstly we give the hyperparameters common to all algorithms (SAC Lag, WCSAC, ORAC) 
\begin{table}[h]
\caption{Common Hyper-Paramters}
\label{tab:EnvHyp}
\vskip 0.1in
\begin{center}
\begin{small}
\begin{sc}
\begin{adjustbox}{width=0.45\textwidth}
    \begin{tabular}{lc}
        \toprule
        \textbf{Common Hyper-parameters} & Value \\
        \midrule
        Policy Lr & $3 \times 10^{-4}$ \\
        Reward Critic Lr & $3 \times 10^{-4}$ \\
        Cost Critic Lr & $3 \times 10^{-4}$ \\
        Agent Entropy AutoTune & True \\
        Agent Entropy Lr & $5 \times 10^{-4}$ \\
        Tau & 0.005 \\
        Target Net Update Frequency & 2 \\
        Buffer Size & $10^6$ \\
        Batch Size & 256 \\
        \bottomrule
    \end{tabular}
    \end{adjustbox}
\end{sc}
\end{small}
\end{center}
\vskip -0.15in
\end{table}

\begin{table}[h]
\caption{Environment Specific Hyper-Paramters}
\label{tab:EnvHyp}
\vskip 0.1in
\begin{center}
\begin{small}
\begin{sc}
\begin{adjustbox}{width=0.8\textwidth}
    \begin{tabular}{lccc}
        \toprule
        \textbf{Hyper-Parameter} & GuardedMaze & SafetyGymnasium & CityLearn\\
        \midrule
        Lagrangian Initial Value & 0.0 & 1.0 & 0.0 \\
        Lagrangian Lr & $5 \times 10^{-4}$&$5 \times 10^{-4}$ &$5 \times 10^{-4}$\\
        Learning Starts & 5000&500&500 \\
        Gamma & 0.9999 &0.99&0.99\\
        Cost Gamma & 0.9999&0.99&0.99 \\
        Policy Hidden Sizes & [64, 64] &[256,256]&[256,256]\\
        Critic Hidden Sizes & [64, 64] &[256,256]&[256,256]\\
        Quantile Hidden Sizes & [64, 64]&[256,256]&[256,256] \\
        Agent Embedding Dim & 64&256&256 \\
        Number of Quantiles (Critic) & 32&32&32 \\
        Number of Quantiles (Policy) & 32&32&32 \\
        Layer Normalisation & True&False&False \\
        Cost Function Ensemble size & 2 & 5 & 5\\
        Penalty Parameter &N/A&10&10\\
        \midrule
        ORAC Explore Delta & 4.0 & 4.0 &4.0 \\  
        ORAC Exploration Reward Beta  & 3.0 & 4.0 & 4.0 \\
        ORAC Exploration Cost Beta  & 2.0 &1 &1 \\
        \bottomrule
    \end{tabular}
    \end{adjustbox}
\end{sc}
\end{small}
\end{center}
\vskip -0.15in
\end{table}

\section{Practical Implementation}
There are a number of practical implementations possible to learn an approximation of the distribution of cost cost returns under policy $\pi$. We detail our uses below
\subsection{Implicit Quantile Networks}
Recent advances in Distributional RL (DRL) allow for the approximation of the full quantile function for the state-action return distribution, using the Implicit Quantile Network (IQN) \citep{dabney_implicit_2018}. 

Letting $F_{Z^{\pi}}^{-1}$ denote the quantile function of the random variable $Z^{\pi}(s,a) = \sum_{t=0}^{\infty}\gamma^{t}\mathrm{C}(s_{t},a_{t},s_{s+1})$, the discounted sum of cost returns under policy $\pi$, we learn a parametrised quantile function of $Z^{\pi}(s,a)$,  mapping the quantile fraction $\tau \in [0,1]$ to the quantile value $F_{Z^{\pi}}^{-1}(\tau)$. For notational simplicity, we write $Z^{\pi}(\tau) := F_{Z^{\pi}}^{-1}(\tau)$. Following \citet{ma_dsac_2020}, we let $\{\tau_{i}\}_{i=0, \cdots, N}$ denote a set of N quantile fractions such that $\tau_{0} = 0,\tau_{i} = \epsilon_{i}/\sum_{i=0}^{N-1}\epsilon_{i}$ where $\epsilon_{i} \sim U[0,1]$. This set of quantile fractions is then ordered such that $\tau_{0}=0$ and $\tau_{N} = 1$ , $\tau_{i} < \tau_{i+1}$, $\forall i < N$ and $\tau_{i} \in [0,1]$. We further denote $\hat{\tau_{i}} = \frac{\tau_{i} + \tau_{i+1}}{2}$.

Using quantile regression, we train the parametrised quantile critic $Z^{\pi}(s,a,\tau|\theta)$ by minimising the weighted pairwise Huber regression loss of different quantile fractions \citep{ma_dsac_2020}. The Huber loss \citep{huber_robust_1964} is defined:
\begin{align}
    \mathcal{L}_\kappa(\delta) = \begin{cases}
                                     \frac{1}{2} \delta^2\quad                   & \text{if } |\delta| \le \kappa \\
                                     \kappa(|\delta| - \frac{1}{2}\kappa)\quad   & \text{otherwise}
                                 \end{cases}
\end{align}
The Huber quantile regression loss is then:
\begin{equation}
    \rho_{\tau}^{\kappa}(\delta_{ij}) = |\tau - \mathbb{I}\{\delta_{ij} <0 \}| \frac{\mathcal{L}_{\kappa}(\delta_{ij})}{\kappa}
\end{equation}
where $\delta_{ij}$ is the pairwise temporal difference error between two quantile fractions $\hat{\tau}_{i}$ and $\hat{\tau}_{j}$ defined:
\begin{equation}
    \delta^{t}_{ij} = c_{t} + \gamma \bigg [ Z^{\pi}(s_{t+1},a_{t+1},\hat{\tau}_{i}|\bar{\theta})\bigg] - Z^{\pi}(s_{t},a_{t},\hat{\tau}_{j}|\theta)
\end{equation}

The learning objective of the parameterised quantile critic $Z^{\pi}(s,a,\tau;\theta)$ is then:
\begin{equation}
    J_{Z}(\theta) = \sum_{i=0}^{N-1}\sum_{j=0}^{N-1}(\tau_{i+1}-\tau_{i})\rho_{\hat{\tau}_{j}}(\delta_{ij}^{t})
\end{equation}

Using this parameterised quantile critic, we estimate the state-action q-value for the cost returns by the expectation over $N$ quantile values:
\begin{equation}
    Q_{C}^{\pi}(s_{t},a_{t},\tau;\theta) = \sum_{i=0}^{N-1}(\tau_{i+1} - \tau_{i})Z^{\pi}(s_{t},a_{t},\hat{\tau}_{i};\theta)
\end{equation}
where $\tau_{i} \sim U[0,1]$, leading to a risk-neutral expectation over the state-action q-value. 
 To induce risk aversion in our agent, we follow the approach laid out by \citet{yang_safetyconstrained_2023} by distorting the base quantile fraction sampling distribution, $\tau \sim U[0,1]$, through the distortion operator $\alpha(\tau)$, such that $\tau \sim U[\alpha, 1]$. This samples quantile fractions from the upper $\alpha$-quantiles of the cost return distribution, allowing us to estimate the CVaR of the policies cost-returns $Z^{\pi}(\tau)$, The CVaR our parametrised value function for a given state-action pair, $\text{CVaR}_{\alpha}\big[Q_{C}^{\pi}(s,a,\tau;\theta)\big]$, is then:
\begin{equation}
\sum_{i=0}^{N-1}(\tau_{i+1} - \tau_{i})\big[Z_{\alpha(\tau)}^{\pi}(s_{t},a_{t},\hat{\tau_{i}};\theta)\big]
\end{equation}
\subsubsection{Optimistic Lower Bound}
To calculate the optimistic lower confidence bound of the state-action cost described in Section \ref{sec:ConfBounds}, we first calculate a lower confidence bound of each quantile value sampled from the quantile function. For each quantile value we calculate its optimistic lower bound by:
\begin{equation}
    \hat{q}_{C}(s,a,\tau_{i}) = \mu_{q_{C}}(s,a,\tau_{i}) - \beta_{C}[\sigma_{q_{C}}(s,a,\tau_{i})]
\end{equation}
where $\mu_{q_{C}}(s,a,\tau_{i}) = \frac{1}{E}\big[\sum_{e=1}^{E} q^{e}_{C}(s,a,\tau_{i})\big]$ is the mean value of the $i$-th quantile values drawn from each critic, and $\sigma_{q_{C}}(s,a,\tau_{i}) = \sqrt{\frac{1}{E}\sum_{e=1}^{E}\big(q^{e}_{C}(s,a,\tau_{i}) -\mu_{q_{C}}(s,a,\tau_{i})\big)^2}$ is the standard deviation of each $i$-th quantile value from the mean quantile value. We then calculate the state-action lower confidence bound as the expectation:
\begin{equation}\label{eq:OptimQsa}
    \hat{Q}_{C,\alpha}^{\pi}(s,a,\tau) = \sum_{i=0}^{N-1}(\tau_{i+1} - \tau_{i}) \; \hat{q}_{C,\alpha}^{\pi}(s,a,\tau_{i})
\end{equation}

\subsection{Spectral Risk - Linear Approximation}
\citet{kim2024spectralrisk} proposed a general approach to linearising spectral risk-measures, such as CVaR, using the dual form of the risk measure, as the standard form of CVaR, such as that approximated using IQN's, is non-linear making it challenging to ensure convergence to optimal policies, particularly in complex environments.
Recall the standard form of a spectral risk-measure of $Z^{\pi}$:
\begin{equation}\label{eq:SpectralRiskFunc}
    \mathcal{R}_{\sigma}(X) \vcentcolon= \int_{0}^{1}F_{X}^{-1}(u)\sigma(u)du
\end{equation}
where $\sigma$ is an increasing function $\sigma \geq 0, \int_{0}^{1}\sigma(u)du = 1$. 
The dual form of E.q. \ref{eq:SpectralRiskFunc} can be expressed by:
\begin{equation}\label{eq:DualSpec}
    \mathcal{R}_{\sigma}(X)= \inf_{g} \mathrm{E}[g(X)] + \int^{1}_{0}g^{*}(\sigma(u))du
\end{equation}
where $g:\mathbb{R} \rightarrow \mathbb{R}$ is an increasing convex functions and $g^*$ is its convex conjugate. CVaR can then be expressed as:
\begin{equation}\label{eq:DualCvar}
    \text{CVaR}_\alpha(X)=\inf_{\beta}\mathbb{E}\bigg[\frac{1}{\alpha}(X-\beta)_{+}\bigg] +\beta
\end{equation}
Here, the integral $g^*(u)$ becomes a constant value of $\beta$ and is independent of $X$, allowing $\text{CVaR}_{\alpha}(X)$ to be expressed using an expectation \citep{kim2024spectralrisk}. 
To practically implement an approximation of CVaR in this form we follow the approach laid out by \citep{kim2024spectralrisk} of discretising the spectral function following:
\begin{equation}
    \sigma(u)\approx\hat{\sigma}(u)\vcentcolon=\eta_{1} +\sum_{j=1}^{M-1}(\eta_{j+1}-\eta_{j})1_{u\geq\alpha_{j}}
\end{equation}
where $0\leq\eta_{j}\leq\eta_{j+1}$, $0\leq\alpha_{j}\leq\alpha_{j+1}\leq1$, and M is the number of discretions. In the case of CVaR, it is already a element of the discretised class of spectral risk measures so can be precisely expressed M=1, by finding the unique values of $1_{u\geq \alpha}/(1-\alpha)$. For example, $\alpha=0.5$ leads to $\eta_{1}=0,\; \eta_{2}= 2$. Using this we parametrise the function $g$ in Eq. \ref{eq:DualSpec} using the parameter $\beta$:
\begin{equation}\label{eq:g(x)}
    g_{\beta}(x)\vcentcolon=\eta_{1}(x) + (\eta_{2}-\eta_{1})(x-\beta)_{+}
\end{equation}
We can now approximate E.q.\ref{eq:DualSpec} with 
\begin{equation}\label{eq:SpecRiskFinal}
    \mathcal{R}_{\hat{\sigma}}^{\beta}\vcentcolon=\mathbb{E}[g_{\beta}(X)] + \int_{0}^{1}g^{*}(\hat{\sigma}(u))du
\end{equation}

This parametrisation presented by \citep{kim2024spectralrisk} allows for a linear approximation of risk, as we can take an expectation over $g_{\beta}(X)$ and calculate a constant value for the integral of the conjugate function. That leaves the problem of learning the optimal value for $\beta$. \citep{kim2024spectralrisk} propose a bilevel optimisation structure for Risk-Averse Constrained RL, where the inner problem finds the optimal policy for the original risk-averse constrained RL problem, with the policy and value functions conditioned on $\beta$:
\begin{equation}\label{eq:WCSACObjective_w_spec}
    \pi(\cdot|\beta)^{*} = \min_{\lambda \geq 0} \max_{\pi} \mathrm{E_{\tau \sim \pi}}[Q^{\pi}_{R}(s,a|\beta) - \lambda ([Q_{C,\alpha}^{\pi}(s,a|\beta)] - c)]
\end{equation}

The outer problem then updates a parametrised distribution, for example a truncated normal distribution, from which $\beta$ is sampled, to find the optimal value of $\beta$ for the following objective:
\begin{equation}
    J(\pi|\beta) = \mathrm{E}\bigg[Q_{R}(s,a|\beta)-K\big(Q_{C,a}(s,a|\beta) - c\big)_{+}\bigg]
\end{equation}

We refer the reader to the excellent paper by \citet{kim2024spectralrisk} for a more detailed discussion of their approach, and its application to a wider spectrum of spectral risk measures.

\subsubsection{Optimistic Lower Bound}
We use Quantile Networks \cite{dabney_distributional_2018} to approximate the full distribution of $Z^{\pi}(s,a)$, and we then calculate \ref{eq:g(x)} by drawing samples from this approximated distribution across K quantiles. To calculate the optimistic lower bound, described in Section \ref{sec:ConfBounds}, we first calculate a lower confidence bound of each quantile value across E parametrised quantile functions. Specifically, for each quantile index \(k \in \{1, \dots, K\}\), we construct the lower confidence bound of the \(k\)-th quantile value as:
\begin{equation}
    \hat{q}_{C}^{k}(s,a) = \mu_{q_{C}}^{k}(s,a) - \beta_{C}[\sigma_{q_{C}}^{k}(s,a)]
\end{equation}
We then calculate the optimistic lower bound of the CVaR by plugging this into e.q. \ref{eq:SpecRiskFinal}
\begin{equation}
    \hat{R}_{\sigma}^{\beta} = \frac{1}{K}\sum_{k=0}^{K}\bigg[\eta_{1}\hat{q}_{C}^{k}(s,a) + (\eta_{2}-\eta_{1})(\hat{q}_{C}^{k}(s,a)-\beta)_{+}\bigg] +  \int_{0}^{1}g^{*}(\hat{\sigma}(u))du
\end{equation}

\section{Comparison of IQN vs Spectral Risk - Linear Approximation}
In this section we present the comparison between the two approximations of CVaR we use in this paper, IQN \cite{dabney_implicit_2018} and linear approximations of Spectral Risk measures \cite{kim2024spectralrisk}, both discussed above. We compare using these two approaches with the WCSAC algorithm, in the Safety Gymnasium environments PointGoal1 and PointButton1. Results are presented across 5 seeds and the agents are trained at $\text{CVaR}_{0.5}$ risk-level. We denote WCSAC using the IQN approximation WCSAC-IQN, and WCSAC using the linear approximation WCSAC-Lin. It is clear from this, as well as the results presented by \cite{kim2024spectralrisk}, that the original approximation of risk used by WCSAC, IQN, fails to satisfy the risk based cost constraints. We reason that this due to the non-linear nature of the IQN approximation of risk and the difficulties that arise in optimising non-linear measures

\begin{table*}[h!]
\caption{Results in SafetyGymnasium's PointGoal1 and PointButton1. Safety threshold set to 10 for all environments.}
\label{tab:SafetyGymWCSACResults}
\begin{center}
\begin{adjustbox}{width=1\textwidth}
\begin{tabular}{lcccccccc}\toprule
    & \multicolumn{4}{c}{PointGoal1} & \multicolumn{4}{c}{PointButton1}  \\
    \cmidrule(lr){2-5} \cmidrule(lr){6-9} 
    & Reward & Mean Cost & $\text{CVaR}_{0.5}$ & $\text{CVaR}_{0.25}$  & Reward & Mean Cost & $\text{CVaR}_{0.5}$ & $\text{CVaR}_{0.25}$  \\
    \midrule
    WCSAC-IQN               &34.9$\pm$0.5            &8.0$\pm$2.6          &14.3$\pm$4.2            &19.7$\pm$6.4
                            &16.2$\pm$4.9            &14.8$\pm$3.4         &25.8$\pm$5.7            &35.9$\pm$7.0 \\
                            
    WCSAC-Lin               &33.9$\pm$0.5           &5.1$\pm$1.0           &\textbf{9.0$\pm$1.6}            &12.3$\pm$1.9
                            &14.6$\pm$3.5           &7.2$\pm$2.5         &\textbf{13.0$\pm$4.4}           &17.6$\pm$5.3 \\

    \bottomrule
\end{tabular}
\end{adjustbox}
\end{center}
\end{table*}

\newpage
\section{Ablations}
ORAC introduces a small number hyperparameters that each affect its performance. To assess the impact of each value we train ORAC under various hyperparameter values, namely the $\beta$ values used to calculate the upper and lower bounds on the reward and cost function, respectively, and the size of the cost function ensemble. We train ORAC in the PointButton environment at CVaR 0.5, following the main experiments, varying the value for each of these hyperparameters and assess their respective impact on the performance on ORAC.
\subsection{Optimistic Lower Bound - Cost Function}
We begin by establishing the impact of varying the value for the optimistic lower bound, $\beta_{C}$, of the state-action cost quantile function. Recall that the locally optimistic lower bound state-action cost value is calculated through Eq. \ref{eq:Optim_qsa}:
\begin{equation}
    \hat{q}_{C}^{k}(s,a) = \mu_{q_{C}}^{k}(s,a) - \beta_{C}[\sigma_{q_{C}}^{k}(s,a)]
\end{equation}
Table \ref{tab:BetaC_Ablation} outlines ablations across 3 values of $\beta_{C}$, results for value 0.5 and 1.5 are reported to 4 seeds whilst results for the value 1.0 are obtained from the main set of experiments.
We note that the value used in the main set of experiments, 1, finds the best constraint satisfying solution. Interestingly the lowest value of  $\beta_{C}$, 0.5, higher reward than the other values but also achieved higher cost. This might indicate that there is a threshold value, above which optimistic exploration improves the policy's cost performance by gathering more useful experience that the policy is uncertain about. 
\begin{table*}[h!]
\caption{Results in SafetyGymnasium's PointButton1, across different values for $\beta_{C}$. Safety threshold set to 10 for all environments.}
\label{tab:BetaC_Ablation}
\begin{center}
\begin{adjustbox}{width=0.5\textwidth}
\begin{tabular}{lccccc}\toprule
    &&  \multicolumn{4}{c}{PointButton1 $\text{CVaR}_{0.5}$}\\
    \cmidrule(lr){3-6}
    &$\beta_{C}$ & Reward & Mean Cost & $\text{CVaR}_{0.5}$ & $\text{CVaR}_{0.25}$ \\
    \midrule
    WCSAC &             &14.2$\pm$3.5  &6.7$\pm$2.4   &11.9$\pm$2.8      &16.5$\pm$6.0 \\
    \midrule
    \multirow{4}{*}{ORAC} & 0.5 
                        &14.0$\pm$2.2           &7.8$\pm$2.4         &14.3$\pm$4.3 &20.4$\pm$7.2\\ 
    \cmidrule{2-6}
    & \textbf{1} &13.9$\pm$2.7&5.4$\pm$2.1          &9.6$\pm$3.5             &13.2$\pm$5.2\\
    \cmidrule{2-6}
    & 1.5              &13.8$\pm$2.6           &5.9$\pm$3.9             &10.3$\pm$6.7 &15.0$\pm$9.4 \\   
    \bottomrule
\end{tabular}
\end{adjustbox}
\end{center}
\end{table*}
\subsection{Optimistic Upper Bound - Reward Function}
Next we evaluate the impact of the optimistic uppper bound value $\beta_{R}$ of the state-action reward function. Recall that the locally optimistic upper bound state-action reward value is calculated through Eq. \ref{eq:OptimQ_r}:
\begin{equation}
    \hat{Q}_{R}^{\pi}(s,a) = \mu_{Q_{R}}(s,a) + \beta_{R}[\sigma_{Q_{R}}(s,a)]
\end{equation}
Table \ref{tab:BetaR_Ablation} outlines ablations across 3 values of $\beta_{R}$, results for value 2 and 5 are reported to 4 seeds whilst results for the value 4 are obtained from the main set of experiments.
We note again the value, 4, used in the main set of experiments finds the best solution, satisfying the constraint whilst only slightly reducing reward compared to WCSAC. The value 2 shows lower reward than the value 4, indicating ORAC is not optimistic enough in terms of reward. The value 5 improves slightly in terms of reward over the value 4, but achieves higher cost. This trend indicates similarly  to the lower bound value $\beta_{C}$, that there is a sweet spot value that maximises reward whilst satisfying the cost constraint. 

\begin{table*}[h!]
\caption{Results in SafetyGymnasium's PointButton1, across different values of $\beta_{R}$. Safety threshold set to 10 for all environments.}
\label{tab:BetaR_Ablation}
\begin{center}
\begin{adjustbox}{width=0.5\textwidth}
\begin{tabular}{lccccc}\toprule
    &&  \multicolumn{4}{c}{PointButton1 $\text{CVaR}_{0.5}$}   \\
    \cmidrule(lr){3-6}
    &$\beta_{R}$ & Reward & Mean Cost & $\text{CVaR}_{0.5}$ & $\text{CVaR}_{0.25}$  \\
    \midrule
    WCSAC &             &14.2$\pm$3.5  &6.7$\pm$2.4   &11.9$\pm$2.8      &16.5$\pm$6.0 \\
    \midrule
    ORAC & 2&13.6$\pm$1.1           &6.2$\pm$0.7         &11.7$\pm$1.5 &18.1$\pm$3.8 \\       
     \cmidrule{2-6}
    & \textbf{4}             &13.9$\pm$2.7&5.4$\pm$2.1          &9.6$\pm$3.5             &13.2$\pm$5.2\\    
    \cmidrule{2-6}
    & 5              &14.0$\pm$1.5           &7.8$\pm$1.3             &13.5$\pm$1.8 &17.0$\pm$2.9   \\  
    \bottomrule
\end{tabular}
\end{adjustbox}
\end{center}
\end{table*}

\newpage
\subsection{Cost Function Ensemble Size}
To evaluate the performance impact of increasing the number of cost functions that make up the ensemble used to train both the baseline WCSAC and ORAC, we vary the size of the ensemble and detail the changes in episodic reward and cost, and CVaR costs. We trained 4 seeds of each algorithm is trained for 5 million timesteps. 

ORAC outperforms WCSAC in terms of reward with an ensemble larger than 2 cost functions in the more challenging CVaR 0.25 experiments. When paired with an ensemble of 8 cost functions, ORAC performs better than the main experimental setup of in the challenging CVaR 0.25, but worse in the CVaR 0.5 setting. Otherwise, WCSAC and ORAC with 5 ensembles performs the best consistently across the two experiment settings. 
\begin{table*}[h!]
\caption{Results in SafetyGymnasium's PointButton1, across different cost function ensemble sizes. Safety threshold set to 10 for all environments.}
\label{tab:EnsembleSize}
\begin{center}
\begin{adjustbox}{width=1\textwidth}
\begin{tabular}{lccccccccc}\toprule
    &&  \multicolumn{4}{c}{PointButton1 $\text{CVaR}_{0.5}$} & \multicolumn{4}{c}{PointButton1 $\text{CVaR}_{0.25}$}  \\
    \cmidrule(lr){3-6} \cmidrule(lr){7-10} 
    &No. of Cost Functions & Reward & Mean Cost & $\text{CVaR}_{0.5}$ & $\text{CVaR}_{0.25}$  & Reward & Mean Cost & $\text{CVaR}_{0.5}$ & $\text{CVaR}_{0.25}$  \\
    \midrule
    WCSAC &\multirow{2}{*}{2}   &13.8$\pm$2.5  &8.6$\pm$2.2         &15.5$\pm$4.0 &22.5$\pm$5.3                                                                           &11.8$\pm$2.9    &5.5$\pm$1.2     &10.4$\pm$2.0 &16.4$\pm$3.9  \\
    
    ORAC &                      &15.3$\pm$3.8   &5.7$\pm$2.9   &9.9$\pm$4.6  &14.1$\pm$6.6
                                &9.8$\pm$ 3.6  &4.0$\pm$2.5   &7.5$\pm$4.6 &11.3$\pm$8.4  \\ 
                               
    \midrule
    WCSAC &\multirow{2}{*}{3}   &13.8$\pm$8.2            &6.1$\pm$1.7         &10.8$\pm$4.0   &15.4$\pm$4.4            
                                &6.6$\pm$2.1           &3.0$\pm$0.8            &5.5$\pm$1.5   &7.7$\pm$2.4  \\
                            
    ORAC &                      &14.5$\pm$2.3           &8.3$\pm$2.0          &14.6$\pm$4.0            &19.1$\pm$4.6
                                &8.4$\pm$4.8           &4.7$\pm$1.8            &8.3$\pm$3.3 &12.9$\pm$5.8   \\ 
    \midrule
    WCSAC &\multirow{2}{*}{\textbf{5}}
                                &14.2$\pm$3.5           &6.7$\pm$2.4            &11.9$\pm$2.8           &16.5$\pm$6.0          
                                &10.1$\pm$3.7           &4.7$\pm$1.1            &8.3$\pm$1.5            &11.4$\pm$2.8\\
                            
    ORAC &                      &13.9$\pm$2.7             &5.4$\pm$2.1          &9.6$\pm$3.5             &13.2$\pm$5.2
                                &11.2$\pm$3.1   &4.7$\pm$1.8   &8.7$\pm$3.0  &12.7$\pm$4.7  \\ 
    \midrule
    WCSAC &\multirow{2}{*}{8}   &14.1$\pm$2.5            &8.9$\pm$7.8          &16.0$\pm$14.5          &26.1$\pm$28.2
                                &6.7$\pm$4.2           &4.9$\pm$0.7            &8.4$\pm$0.9 &12.0$\pm$1.8 \\
                            
    ORAC &                      &13.8$\pm$1.7           &7.8$\pm$1.7            &14.1$\pm$5.2   &21.5$\pm$5.8
                                &12.7$\pm$4.1           &4.0$\pm$1.7            &6.4$\pm$2.2    &8.2$\pm$2.5  \\

    \bottomrule
\end{tabular}
\end{adjustbox}
\end{center}
\end{table*}
\end{document}